\documentclass[letterpaper,twocolumn,10pt]{article}
\usepackage{usenix-2020-09}

\usepackage{tikz}
\usepackage{amsmath}

\usepackage{tikz}
\usepackage{xcolor}
\usepackage{listings}
\usepackage{float}
\usepackage{svg}
\usepackage{pdfpages}
\usepackage{graphicx}
\usepackage{caption}
\usepackage{subcaption}
\usepackage{tabularx}
\usepackage{booktabs}
\usepackage{algorithm}
\usepackage{algpseudocode}

\newcommand{\proj}{\texttt{MOSEL}}

\newcommand{\ballnumber}[1]{\tikz[baseline=(myanchor.base)] \node[circle,fill=.,inner sep=1pt] (myanchor) {\color{-.}\bfseries\footnotesize #1};}

\usepackage{filecontents}

\begin{document}

\date{}

\title{\Large \bf \proj{}: Inference Serving Using Dynamic Modality Selection}
\author{Bodun Hu\qquad Le Xu\qquad Jeongyoon Moon\qquad Neeraja J. Yadwadkar\qquad Aditya Akella\\UT Austin}

\maketitle

\begin{abstract}
Rapid advancements over the years have helped machine learning models reach previously hard-to-achieve goals, sometimes even exceeding human capabilities. However, to attain the desired accuracy, the model sizes and in turn their computational requirements have increased drastically. Thus, serving predictions from these models to meet any target latency and cost requirements of applications remains a key challenge, despite recent work in building inference-serving systems as well as algorithmic approaches that dynamically adapt models based on inputs.
In this paper, we introduce a form of {\em dynamism}, modality selection, where we adaptively choose modalities from inference inputs while maintaining the model quality. We introduce \proj{}, an automated inference serving system for multi-modal ML models that carefully picks input modalities
per request based on user-defined performance and accuracy requirements. \proj{} exploits modality configurations extensively, improving system throughput by 3.6$\times$ with an accuracy guarantee and shortening job completion times by 11$\times$. 

    
\end{abstract}





\section{Introduction}


With recent advancements in Deep Learning Models (DNNs), especially Transformers, Machine Learning (ML) has been far exceeding human capabilities in various Computer Vision and Natural Language Processing tasks~\cite{kaimingrectifier,wolf2019huggingface}.
However, the computational requirement of the largest ML models/applications has doubled every few months, resulting in a 1,000,000$\times$ increase from 2012 to 2020~\cite{visparams}. The increasing size of the models presents fundamental challenges in terms of latency and cost when they are commissioned for inference~\cite{infaas, cocktail,clockwork}.

To combat the overheads associated with model sizes, 
today's services often
replace a large model with a single, cheaper variant, typically obtained using ML \emph{compression techniques}, such as distillation~\cite{distilbert, mullapudi2019online}, pruning~\cite{runtimepruning,gordon-etal-2020-compressing} and quantization~\cite{polino2018model}.  Although compression techniques are quite successful in achieving their design goals in small and medium-sized models, larger and more sophisticated models (e.g., autoregressive/generative models) face new challenges: it may not always be
possible to produce a small enough model that matches (or is close to) the
original model in accuracy while meeting the cost constraints for all inputs. 

Alternatively, approaches that adapt the models dynamically based on the inputs have been proposed. 
Such techniques that explore dynamic adaptation opportunities of a model fall into two classes: (i) techniques that reuse the original model and (ii) techniques that rethink the original model. In the former, the key idea is to maintain the model's original architecture and augment it with adaptation capabilities. Examples of these include early-exiting~\cite{deebert, pabee,branchynet, berxit}, and layer-skipping~\cite{tang2022tvlt}. 
The latter techniques modify and/or replace the original model with equivalents that support adaptation; examples in this category include ensembling~\cite{power-of-ensembles}, model merging~\cite{gemel}, and mixture-of-experts~\cite{megablocks,deepspeedmoe}. 
Such dynamism offers latency and resource savings~\cite{branchynet,berxit,beluch2018power}, but demands new systems designs that are both adaptive and flexible.  

In this paper, we introduce a new form of dynamism that falls into the former category. In particular, we propose {\em modulating the input}, specifically via {\em selectively using} parts of it. We develop this idea and demonstrate its usefulness in the context of {\em multi-modal learning}~\cite{ngiam2011multimodal, baltruvsaitis2018multimodal}, an emerging and important class of ML techniques that combine data of different modalities to
provide prediction cooperatively, enhancing ML models’
prediction accuracy. 

As we describe in Section~\ref{sec:observation}, we empirically find that, for inference using such models, some modalities (e.g., the audio modality in the Textless
Vision-Language Transformer (TVLT) model~\cite{tang2022tvlt}) contribute
significantly to prediction accuracy while not being the major contributor to resource use (e.g., memory) and processing time, while
other modalities (e.g., the video modality in TVLT) consume a significant
amount of resources and incur latency while only marginally contributing to
prediction accuracy. The observations depend on the inputs being processed and the type of model in use. 
We leverage 
the above insight in inference settings and 
propose that modalities be selectively enabled or disabled in a given sequence of inference requests, depending on application requirements and workload patterns and distributions, creating novel opportunities to exploit the trade-off between memory/speed and accuracy that multi-modality presents. We refer to this as {\em modality selection}, and observe that it complements the first set of dynamism techniques above that modulate or change the model being deployed. It can also be directly applied to the original model, and it is complementary to compression and distillation techniques. 

We build \proj{}, an automated inference system for multi-modal models that carefully preforms input modalities selection per request based on user-defined performance and accuracy requirements. We develop \proj{} to ensure scaling and performance at inference time by dividing it into offline and online portions. For the former, we develop novel offline profiling strategies that, given a batch of inference requests from a job, help quickly determine at run-time what modalities to use for each request so as to meet the latency requirement at a suitable accuracy. For the latter, we develop dynamic techniques to ensure that, at inference time, late-enqueued jobs don't miss their latency requirements. We do this by facilitating jobs ahead in the inference queue to dynamically reselect modalities to ease the queueing load; this allows later jobs to run at the needed accuracy without missing their latency targets. 


We evaluate \proj{} on a set of representative multi-modal models that utilize commonly-seen architectures (Transformer~\cite{vaswani2017attention}, BERT~\cite{devlin2019bert}, CNN~\cite{lecun2015deep}). In our evaluation, \proj{} outperforms \textit{modality-agnostic} approaches in resource utilization and query spike tolerance. \proj{} is capable of reducing job completion time by up to $11\times$ and handling up to $3.6\times$ more requests with accuracy guarantees. Moreover, \proj{} can achieve up to $4.6\times$ throughput when combined with quantization.


\section{Background and Motivation}

\begin{figure*}[thp!]
    \centering
    \includegraphics[width=1\textwidth]{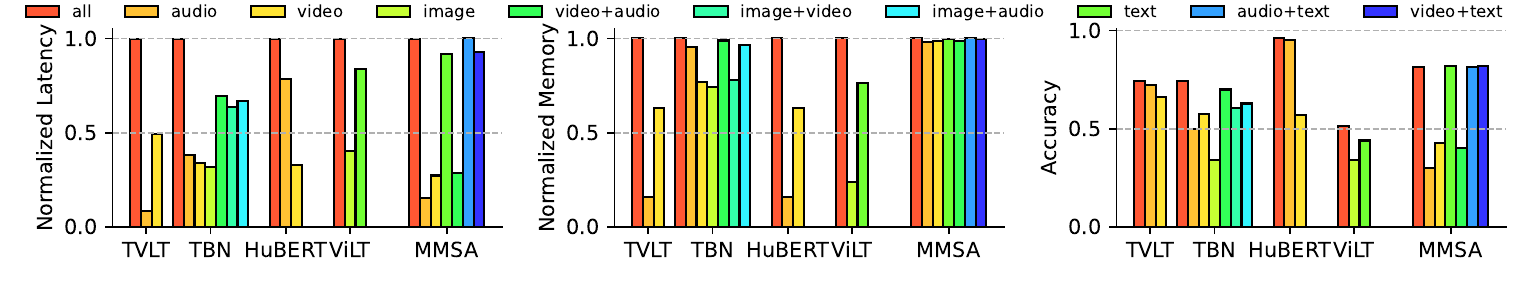}\hfill
    \includegraphics[width=1\textwidth]{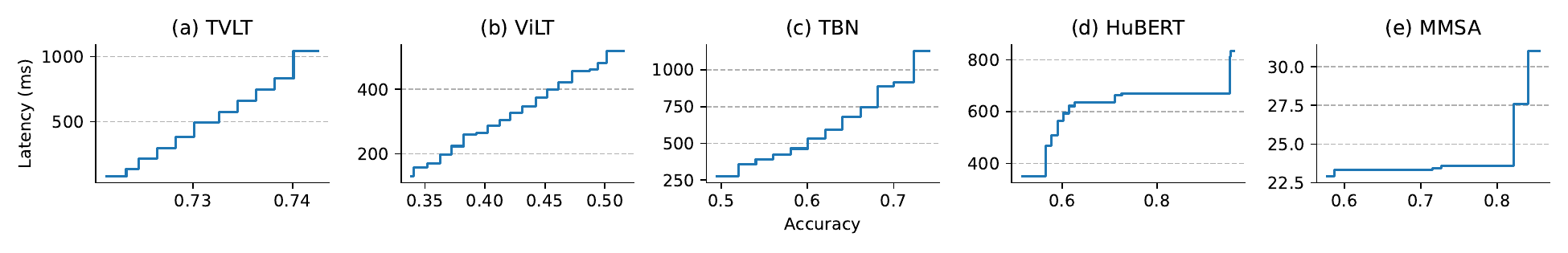}\hfill
    \vspace{-1em} 
    \caption{\textit{Performance comparison of different modalities for models discussed in Table~\ref{tab:model_overview}:
        (Upper Left) The normalized latency for different modalities. (Upper Middle) The normalized memory footprint of different modalities. (Upper Right) Accuracy comparison using different modalities. (Bottom) Minimum latency required to achieve different levels of accuracy across various models using combinations of modalities.}
    }
    \vspace{-1.5em} 
    \label{fig:background}
\end{figure*}

\subsection{Background}
\label{subsec:background}

\noindent\textbf{Inference and its challenges:} Inference systems use pre-trained ML models to make predictions or perform tasks based on input data. With the increasing availability of datasets and progress in ML research, ML models have become more accurate and capable, leading to increasing adoptions in production settings~\cite{hazelwood2018applied, gupta2020architectural}. Inferences tasks have shown to dominate ML production costs:~\cite{romero2021infaas,aws_inf}.


Inference systems typically serve prediction for a variety types of user requests~\cite{crankshaw2017clipper, reddi2020mlperf, romero2021infaas}. For inference systems, accuracy and latency are two key factors that impact end-user experiences~\cite{romero2021infaas}. Based on deployment scenarios, the requirements imposed by different applications one or the other ~\cite{hsieh2018focus} or both~\cite{gog2022d3} of these factors are vastly different~\cite{reddi2020mlperf}. To complicate the issue even further, inference systems often need to provision for dynamic workloads~\cite{yadwadkar2019case,inferline,crankshaw2017clipper,shepherd}, making dynamic resource provisioning another key requirement for cost/resource effectiveness.


Moreover, the above challenges are magnified 
when multiple services leverage the same model for inference tasks. Increasing accuracy for one query exercising a shared model directly impacts the availability of resources for subsequent queries, either delaying their processing or affecting their accuracy. Furthermore, some services may prioritize lower latency to ensure faster response times, and others may prefer good throughput, resulting in varying SLOs.


\noindent\textbf{Multimodal Learning:}
Multimodal learning techniques 
are shown to surpass the abilities of unimodal techniques by exploiting the complementary nature of different modalities, such as text, image, audio, and video, in input data sources~\cite{ngiam2011multimodal, baltruvsaitis2018multimodal}.
Multimodal machine learning techniques rely on ways to fuse data from diverse modalities~\cite{baltruvsaitis2018multimodal}. 
The existing approaches for fusion can be broadly classified into two categories: early fusion~\cite{snoek2005early,Atrey2010MultimodalFF,7194741} and late fusion~\cite{snoek2005early,8578225,8953429}. Early fusion consolidates modalities at an early stage, blending features prior to further processing. For example, TVLT~\cite{tang2022tvlt} converts video and audio data into sequences of patches and concatenates them before the transformer layer. On the other hand, late fusion preserves separate pathways for each modality and merges the outcomes later. For example, Temporal Binding Network (TBN)~\cite{kazakos2019epic} processes RGB, Flow, and audio separately, then combines them with mid-level fusion alongside sparse temporal sampling of fused representations. Some fusion methods attempt to combine properties from both early and late fusion~\cite{NEURIPS2021_76ba9f56, 9156844, 8954353, vielzeuf2018centralnet, xue2023dynamic, NEURIPS2021_76ba9f56}.



%

\subsection{Characteristics of multi-modality ML models}
\label{sec:observation}





We now characterize the trade-offs in multi-modal inference, using them to motivate techniques for achieving better inference latency and resource-efficiency. 

\noindent\textbf{Accuracy Across Modalities:} 
Multimodal deep neural networks (DNNs) can achieve higher accuracy than unimodal models by combining multiple modalities, such as visual, audio, and textual data. However, the relative importance of each modality may depend on various factors, such as the task, the data, and the model architecture. Previous studies~\cite{ma2021smil,9879085,tang2022tvlt,NEURIPS2021_76ba9f56} have shown that different modalities can have different effects on the performance of multimodal models on different validation datasets. Figure~\ref{fig:background} shows the average accuracy of different multimodal models using all possible combinations of modalities. 
We observe that some models (TVLT) are capable of achieving high accuracy without using all the available modalities in input data. 

\noindent\textbf{Performance Implications:}
Different modalities have different data representations and processing methods. 
Thus, along with their impact on model accuracy, different modalities impact the latency and memory consumption of multimodal models differently. 
Figure~\ref{fig:background} compares the inference latency and memory consumption of different models. We observe that for TBN, the video modality is more costly than the other modalities in terms of both time and space, due to the large data size of the temporal dimension. TBN uses late-fusion, and so most of the computation happens before fusion.  This implies that adding more modalities are likely to increase latency and resource consumption.

For the attention-based models such as the TVLT~\cite{tang2022tvlt} (shown in Figure~\ref{fig:background}), we find that the audio modality is more efficient than the video modality in terms of memory usage and processing latency, with only a minor trade-off in accuracy. The memory consumption of TVLT~\cite{tang2022tvlt} depends on the attention mechanism \cite{vaswani2017attention}, which scales quadratically with the sequence length. It concatenates different modalities into one sequence, which grows longer as more modalities are added. Thus, using selected few modalities reduces sequence length which results in reduced overall memory consumption of the model.
Shorter sequence length also means less computation for serving an inference from the model. 
A similar approach of using attention-based multi-modal models is adopted by many recent works~\cite{tang2022tvlt,shi2021learning,NEURIPS2021_76ba9f56,NIPS2016_82b8a343,vilbert,videobert}, which may also achieve lower latency and memory consumption by utilizing fewer modalities.

\noindent\textbf{Opportunities:}
The diverse requirements of applications in terms of the desired accuracy, memory availability, and target latency across different modalities offer interesting possibilities for adaptive multimodal selection, which previous inference-serving systems have not considered. In particular, modalities can be selectively enabled or disabled depending on the application requirements and availability of underlying resources. For instance, for TVLT, as shown in~Figure~\ref{fig:background}, under high load conditions, we can use only the audio modality and achieve a 11$\times$ latency reduction without sacrificing much accuracy. Whereas, when system load is low, both video and audio modalities can be used to obtain the highest accuracy, at the cost of higher latency. Figure~\ref{fig:background} (Bottom) shows the optimal modality choices for minimizing the latency of 32 requests with a given accuracy objective. Section~\ref{sec:optimize-mod-selection} discusses how to generate these modality choices. 

\section{Challenges}
\label{sec:challenges}

Typically, applications provide the inference systems with SLOs such as accuracy and latency. We define a \textit{job} as a set of requests. Each job has a specified accuracy and latency constraint.  \textit{Effective accuracy} is the average accuracy that is achieved across all requests in the job at inference time. This must equal or exceed the job's specified accuracy requirement. Similarly, all requests in the job must be completed within the specified latency. 

\subsection{Challenges in Exploiting Multiple Modalities}
\label{subsec:mod_select}

\begin{figure}[!ht]
    \centering
    \includegraphics[width=\linewidth]{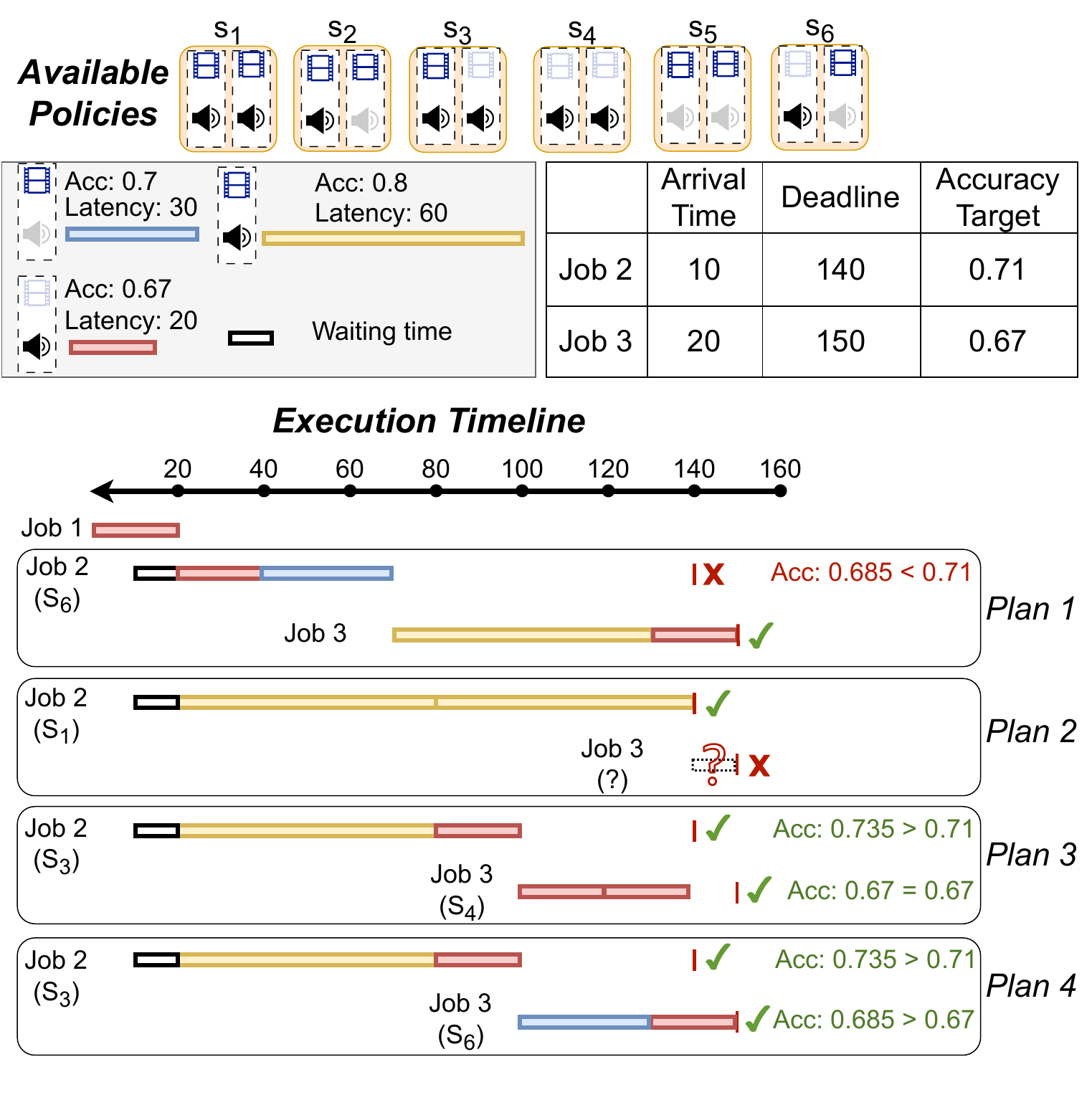}
    \vspace{-2em}
    \caption{\it : (a) Job 1, which has only one audio modality, executes from timestamp 0 to 20. (b) Job 2 arrives at timestamp 10 and executes from timestamp 20, after job 1 completes. One of job 2's modality selection strategy, $s_6$, shown in Plan 1, has an effective accuracy of $\frac{0.67+0.7}{2}=0.685$, making it unable to satisfy the accuracy requirement of job 2 (Plan 1). Similarly, $s_4$ also fails to satisfy the accuracy requirement with an accuracy of $0.67$. (c) Job 3 also arrives at a timestamp a little after 20 and has a deadline constraint of 150. However, if job 2 selects $s_1$, it will occupy the system until timestamp 140, leaving no feasible modality selection strategy for job 3 to meet its deadline (Plan 2). (d) When job 2  selects a modality that yields the lowest accuracy for one of its requests, the leftover system resource could aid job 3 (Plan 3). (e) In fact, the additional resource could be used to further increase the accuracy of job 3 by using the video modality with higher accuracy (Plan 4).
    }
    \label{fig:modality-selection-strategies}
\end{figure}


We illustrate the challenges in exploring multiple modalities using Figure~\ref{fig:modality-selection-strategies}. Here we have three jobs. Job 1 has one request, and Jobs 2 and 3 have two each. Job 1's request only has the audio modality, whereas the other jobs' requests have both modalities. Job 1 runs from time 0 to 20, and Job 2 arrives at time 10. Job 3 arrives soon after. Requests in jobs are executed in a FIFO manner.

\noindent\textbf{Modality search for one job:} To select modalities adaptively, we need a suitable \textit{modality selection strategy} for each job. This is a policy that determines which modalities should be used for each request in a job. Figure~\ref{fig:modality-selection-strategies} illustrates the six possible modality selection strategies for a request in Job 2 or Job 3. For example, in S1, both requests use both modalities, whereas in S4 both requests use only the audio modality. 

The set of strategies can be large, making choosing among them challenging. 
The set grows proportional to the number of requests in a job and exponential in the number of modalities per request. For a job with 20 requests and 3 modalities, there are 231 possible modality selection strategies.  
Also, some of these strategies are infeasible because they may not meet the accuracy and latency constraints of the job. For example, of the six strategies, only two are valid for Job 2, as they satisfy the job's effective accuracy requirement (0.71) without violating the latency constraint (140). 

Therefore, given the large search space of possible strategies, we need efficient methods to prune infeasible strategies and estimate the latency of the feasible ones, especially for jobs with low latency requirements.

\noindent\textbf{Planning for multiple jobs:} 
Figure~\ref{fig:modality-selection-strategies} illustrates that multiple strategies can yield valid effective accuracy. But we note that some strategies that create opportunities for a job potentially come at the cost of other jobs. 
In particular, greedily increasing accuracy for a job comes at the cost of increased resource consumption that may in turn hurt other jobs. This is illustrated by Plan 2 in Figure~\ref{fig:modality-selection-strategies}, which offers great accuracy for Job 2 by selecting both modalities for both requests (effective accuracy of 0.8) and finishing exactly by 140 time units, but it leaves no room for Job 3 to finish by its deadline. On the other hand, by lowering accuracy for some jobs, we are left with extra resources that can be used to improve the outcomes for other jobs; e.g., in Plan 3, we use just the audio modality for one of Job 2's requests, yielding an effective accuracy of 0.735, which allows Job 3 to start at time 100 and use the audio modality for both its requests in order to finish by time 140 with an accuracy of 0.67. In fact, we can do better for Job 3 -- by picking a higher-accuracy modality (video) for one of its requests, Job 3 achieves an effective accuracy of 0.685 (Plan 4), while finishing at its deadline of 150. 

The upshot is that we may have to look for less-than-optimal strategies for some jobs in the queue to enable other later-coming jobs to meet their objectives.

To tackle the challenges for models that dynamically adapt to input data, we need techniques that adapt to the changing SLO requirements and query load across jobs. 
To tackle such dynamics, existing inference serving systems leverage various techniques, including autoscaling~\cite{azure_ml, sagemaker}, model switching~\cite{infaas, zhang2020model}, batching~\cite{clipper}, predictive serving~\cite{clockwork} and preemption~\cite{shepherd}. 
Inference systems for multi-modality such as~\cite{li2021low} focus on speculatively executing modalities. All of these techniques are agnostic to input data modalities and to the possibility of exploiting them for performance and efficiency. 



\section{\texttt{\proj~}Overview}

\subsection{Design Goals}

\begin{figure}[t]
    \centering
    \includegraphics[width=\linewidth]{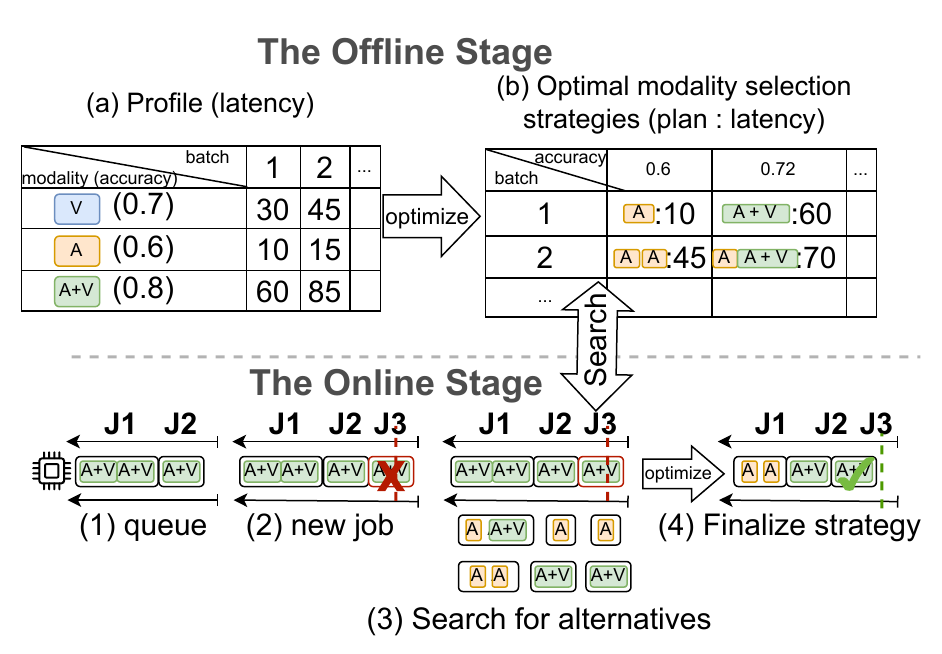}
    \vspace{-2em}
    \caption{
    \textit{\proj{} first construct optimized modality selection strategy matrix based on model profile offline. Then it uses this matrix to dynamically derive modality deployment strategies for different jobs online. }
    }
    \vspace{-2em} 
    \label{fig:optimization-overview}
\end{figure}


We design \proj{} to meet several goals. 
First, \proj{} should automate modality selection: users should not specify modality choices for jobs; they should only focus on high-level performance, costs, or accuracy requirements. Second, to make modality selection practical during deployment, the system should strive to minimize the overhead of modality selection. Third, given the dynamic variation of the system load, we need to track the system state and adjust the modality selection plan accordingly, ensuring that jobs continue to meet their objectives as best as possible. In this section, we give an overview of our approach that meets these goals and addresses the challenges presented in Section~\ref{sec:challenges}.


Figure~\ref{fig:optimization-overview} illustrates the two stages of \proj{}'s approach: \textit{offline profiling} and \textit{online optimization}. We outline these next, and discuss their algorithmic details in Section~\ref{sec:problem-formulation}. 

\subsection{The Offline Stage}
\label{sec:optimize-mod-selection}

As discussed in Section~\ref{subsec:mod_select}, a single job could have various modality selection strategies. Some of these strategies may fail to meet a job's latency goal or SLO. To narrow down the search space, \proj{} needs to find a modality selection strategy with the \textit{low enough} latency, while still satisfying the given accuracy requirement. 

We generate suitable modality selection strategies {\em offline} by constructing {\em profiles} of the accuracy and latency of each modality using validation datasets. 
In particular, we evaluate and record the latency of various combinations of modalities under different batch sizes for a given model. Figure~\ref{fig:optimization-overview}(a) shows the latency for each combination of modalities and batch size as an entry in a table.




Next, \proj{} uses the accuracy and latency measurements to construct a matrix of modality selection strategies for different job sizes and accuracy constraints for a given model, as shown in Fig~\ref{fig:optimization-overview}(b). Each matrix entry shows the best modality combinations for minimizing latency and meeting accuracy needs for a given job size (number of requests), as well as the latency of the chosen combination. The construction is again offline and only happens once before the model is deployed. \proj{} uses a non-linear integer program (NILP) to construct this matrix; more details can be found in Section~\ref{sec:offline_formulation}. 

\subsection{The Online Stage}
\label{sec:search_optimize}

Once a model is deployed, the system queues incoming jobs. Further, \proj{} prioritizes and orders jobs by their deadlines, as shown Figure~\ref{fig:optimization-overview}(1) (bottom half; leftmost panel). Each new job adopts the strategy with the highest accuracy by default.

\proj{} monitors the queued jobs and detects if incoming new jobs may suffer from deadline violations. 
If a job risks missing its deadline, as shown in Figure~\ref{fig:optimization-overview}(2), its preceding jobs need to change their modality selection strategies, potentially sacrificing accuracy, but finishing faster and thus reducing the wait time for the job at risk of deadline violation. 

To aid this, \proj{} looks up the pre-computed modality selection strategy matrix for all possible strategies for each queued job, as shown in Figure~\ref{fig:optimization-overview}(3). Note that we do not consider the resource availability of other jobs when choosing a strategy. We only consider the latency and accuracy constraints of each job, assuming it can start right away without any interference.


At this point, \proj{} reassigns a strategy for each job, avoiding deadline violation and shortening queue delay for all jobs, as shown in Figure~\ref{fig:optimization-overview}(4). If the queue is relatively empty, or contains few jobs, \proj{} will attempt to increase the accuracy for all queued jobs by progressively trying modality strategies with higher accuracy for each of them. The modality strategy reassigning process is formulated as an INLP, details in Section~\ref{sec:online-section}.


\section{Formulation}
\label{sec:problem-formulation}

This section describes our nonlinear integer programming (NILP) formulation that identifies the best modality selection strategies to reduce latency and meet accuracy requirements for different job sizes. Then, we explain how we prevent deadline violations and guarantee a fair accuracy distribution among various jobs.

\subsection{Offline Optimal Strategy Generation}
\label{sec:offline_formulation}

As discussed in ~\ref{sec:observation}, different modalities exhibit varying resource consumption patterns. Consequently, such variation would also persist given different batch sizes. For a given model supporting $n$ modalities and batch size of $b$, we measure the latency for each modality combination and batch size, yielding, in total, $b(2^n - 1)$ results. We denote the results collectively as the set $\mathcal{D}$. $\mathcal{D}_{ij}$ represents the latency using modality combination $i$ with batch size $j$.

To produce the modality selection strategies with the lowest latency for a given job size that satisfies an accuracy SLO $\alpha$, we divide the job into multiple batches, each using a different modality selection strategy from $\mathcal{D}$. We aim to find the modality combinations and the exact number of requests they should cover, jointly denoted as $ \{ \mathcal{I}, \mathcal{J} \}$, such that the total latency:
$$
\sum_{i,j \in \mathcal{I}, \mathcal{J}} D_{ij} 
$$  is minimized, subject to the following constraints:
(\textbf{1}) 
The sum of all requests must be equal to the total number of requests.
(\textbf{2}) 
The average accuracy of all requests using different modality combinations must exceed the user-specified accuracy objective $\alpha$. We use $acc(i)$ to denote the effective average accuracy achieved by a modality selection strategy $i$, and $|\mathcal{R}|$ as the size of a given job. Formally, we have:
\begin{align}
  &\begin{aligned}
    |\mathcal{R}| = \sum_{j \in \mathcal{J}}j
  \end{aligned}\\
  &\begin{aligned}
    \sum_{i,j \in \mathcal{I}, \mathcal{J}} acc(i)j \geq \alpha |\mathcal{R}|
  \end{aligned}
\end{align}
When the target model is deployed, it may receive inference jobs with different sizes and accuracy requirements. Running the NILP entirely online may increase the risk of deadline violations due to the solver’s overhead, especially for inference tasks with strict latency constraints. Since the optimization process depends only on accuracy and profiled latency $\mathcal{D}$, we can generate the optimal modality selection strategies for the most common job sizes and accuracy goals completely offline. Formally, given $N$ possible request sizes and $A$ accuracy requirements, we optimize for each of the $N\cdot A$ combinations. This process has negligible overheads.

\subsection{Online Modality Selections and Adjustment}
\label{sec:online-section}


We formulate the problem of \proj{} selecting alternative modality strategies for jobs in a queue when an incoming job is at risk of deadline violation as an integer non-linear program (INLP), which is a type of optimization problem that involves nonlinear functions and integer variables. 
Section~\ref{sec:evaluation} demonstrates the evaluation setup and the performances of our \texttt{optimized} modality-aware approach.

To ensure that no job in the queue misses its deadline (shown in Figure~\ref{fig:optimization-overview}), we use a variable $T$ to represent the maximum allowed time budget. If \proj{} detects a deadline violation, it sets $T$ to the difference between the violator’s deadline and the start time of the most recent job. Otherwise, it sets $T$ to the difference between the deadline of the last job in the queue and the start time of the most recent job. This means that all jobs before the violator must have a modality selection strategy that results in a total latency lower than $T$.


Each job can have multiple modality selection strategies that satisfy the minimum accuracy objective, as shown in Figure~\ref{fig:optimization-overview}(3). These strategies are precomputed in the offline stage. \proj{} selects one strategy for each job in the queue, depending on whether there is a deadline violation or not. If there is a violator, \proj{} selects one strategy for all jobs before the violator. If no job misses its deadline, \proj{} selects one strategy for all jobs in the queue. We denote the set of all such strategies as $S$. We also denote the set of all jobs before the violator as $J$. The goal is to select a strategy for each job in $J$ that minimizes the total latency and maximizes the total accuracy.

The objective function that our INLP maximizes is the average accuracy for all requests, defined as:
$$\sum_{s, j \in S, J} acc(s) \cdot |j|$$
We use $l(s)$ to represent the execution latency of a strategy. To select a strategy for each job that fits within the total latency budget, we add the following constraint to our INLP: the sum of the execution latency of the selected strategies for all jobs must not exceed the total time budget $T$:
$$
\sum_{s\in S} l(s) \leq T
$$


\begin{algorithm}
\caption{Random modality strategy selection}
\label{alg:random_mod}
\begin{algorithmic}[1]
\Function{RandSelect}{$job, QmodalityStrategies$}
\State $\textit{S} \gets \{\}$
\If{\texttt{hasDeadlineViolation($jobQ$)}}
    \State $J \gets \texttt{jobsBeforeViolator}(jobQ)$
\Else{ }
    \State $J \gets jobQ$
\EndIf
\State $deadline \gets \texttt{overhead}(J) + currentTime$
\While{$deadline > violatorDeadline$}
    \State $j \gets \texttt{randomJob}(J)$
    \State $s \gets \texttt{modalityStrategies}(|j|, $
    \State $\ \ \ \ \ \ \ \ \ \ \ \ \ \texttt{accuracy}(\textit{j}))$
    \State $deadline \gets \texttt{update}(deadline, s)$
    \State $S.\texttt{append}(s)$
\EndWhile
\State \Return{\texttt{S}}
\EndFunction
\end{algorithmic}
\end{algorithm}

\noindent\textbf{A Greedy Heuristic.} The time taken to solve each instance of our INLP is long (up to 70 ms) 
making it challenging to use in an online fashion for jobs with low latency requirements. To address this issue, we also propose a greedy heuristic that adapts the accuracy of enqueued jobs by randomly applying modality selection strategies with the lowest latency and accuracy above the minimum accuracy requirements for jobs before the deadline violator until the queue wait time for the violator is within its deadline. The steps are described in Algorithm~\ref{alg:random_mod}. We present the evaluation setup and the performance of the \texttt{random} heuristic in Section~\ref{sec:evaluation}.


\section{\proj{} Implementation}
\label{sec:implementation}

\proj{} is implemented in 3k lines of Python code. The offline profiler uses Pytorch~\cite{pytorch} to execute  \ballnumber{1} DNNs on the GPU and profile  \ballnumber{2} system metrics through CUDA API. We use GEKKO~\cite{gekko} to generate  \ballnumber{3} the offline modality selection strategies. GEKKO is an optimizer that solves large-scale mixed-integer and differential algebraic equations with nonlinear programming solvers. The generated selection strategies can be stored in a single pickle object. When a model is deployed on an accelerator, a monitor process and a worker process are launched. The monitor process buffers incoming jobs, \ballnumber{4} retrieves the generated modality plans, then \ballnumber{5} uses GEKKO to finalize the modality plan for each job, and puts the job into a FIFO queue shared with the worker process. The GEKKO optimizer can take up 70 ms seconds to generate a solution. Therefore, the monitor process enqueues enough jobs to compensate for the optimizer overhead. In addition, to avoid overflowing the FIFO queue, the monitor process buffers more jobs and pauses periodically. The worker process polls the FIFO queue and executes the \ballnumber{6} the jobs. It also reports the latest \ballnumber{7} execution latency metrics back to the monitor process, so that it can estimate available system resources more accurately.





\begin{figure}[h]
    \centering
    \includegraphics[width=\linewidth]{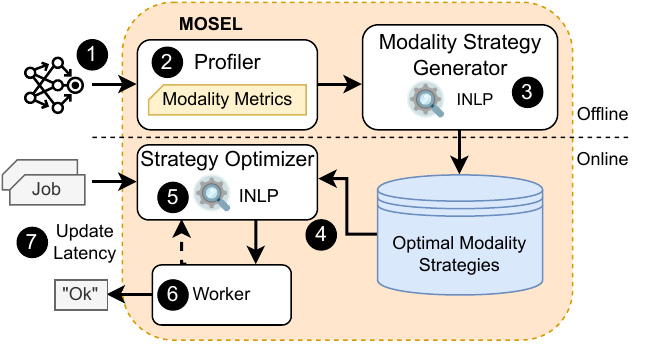}
    \caption{\it \proj{} Workflow}
    \label{fig:placehodler}
    \vspace{-1em} 
\end{figure}

\section{Evaluation}
\label{sec:evaluation}

To evaluate our implementation, we conduct experiments using realistic workloads and address the following questions:

\noindent\textit{{\bf Q1}: What are the benefits of \textit{modality-aware} optimizations?} (Section~\ref{subsec:production})

\noindent\textit{{\bf Q2}: How compatible is \proj{} with existing model optimization techniques? }(Section~\ref{subsec:compatibility})

\noindent\textit{{\bf Q3}: Is \proj{} resilient towards profiling error? }(Section~\ref{subsec:ablation})

Unless specified otherwise, our experiments use the following configurations. 

\noindent\textbf{Experimental Setup} All measurements are conducted on real hardware using a NVIDIA Tesla A100 GPU with 1.48 GHz shader clock and 80GB DRAM. The CPU is an Intel Xeon Silver 4314 running at 2.40GHz (128GiB DRAM). We used NVIDIA driver version 525.85 and CUDA version 12.0. The PyTorch version is 2.1.0. The operating system is Ubuntu 22.04.1 LTS with 5.15.0 kernel.


\noindent\textbf{Models.} Table~\ref{tab:model_overview} summarizes the five models used for evaluation; all are pre-trained using PyTorch. The models differ in size and fusion strategy. All models are finetuned on the task-specific datasets, and preloaded onto the GPU before evaluation.

\noindent\textbf{Workloads.} We conducted experiments using both synthetic and real-world query patterns. For synthetic workloads, we generated queries with constant loads and a fixed time interval. For real-world workloads, we used the timing information from a Twitter trace, collected over a month in 2018~\cite{twitter_trace}. Previous work on inference serving has shown this trace reflects realistic inference workloads with diurnal patterns and unexpected spikes~\cite{234998}. For each experiment, we randomly selected a day out of the month from the Twitter trace.

\begin{table*}[th!]
  \centering
\small
\begin{tabular}{p{0.2\linewidth} |p{0.3\linewidth} |p{0.2\linewidth} |p{0.15\linewidth} |p{0.05\linewidth} }
  \hline
    \textbf{Task} & \textbf{Dataset} & \textbf{Model} & \textbf{Modalities} & \textbf{Fusion} \\
    \hline
    Sentiment Analysis & MOSEI~\cite{zadeh2018multimodal} & TVLT~\cite{tang2022tvlt} & audio, video & Early\\
    \hline
    Speech Recognition & LRS3~\cite{afouras2018lrs3} & AVHuBERT~\cite{shi2022learning} & audio, video & Early\\
    \hline
    Action Recognition & EPIC-KITECHENS~\cite{Damen2022RESCALING} & TBN~\cite{kazakos2019epic} & audio, video, image & Late\\
    \hline
    Sentiment Analysis & MOSEI~\cite{zadeh2018multimodal} & Self-MM~\cite{yu2021learning} & text, audio, video & Late\\
    \hline
    Multi-Label Classification & MM-IMDb~\cite{mmimdb} & ViLT~\cite{kim2021vilt} & text, image & Early\\
    \hline
  \end{tabular}

  \caption{
  \textit{
  Tasks, datasets used for finetuning and evaluation, model architectures, model sizes, modalities used, fusion strategy}
  }
  \vspace{-1em} 
  \label{tab:model_overview}
\end{table*}


\subsection{\proj{} with production workload}
\label{subsec:production}

To show that through dynamic modality selection mechanism, \proj{} improves the throughput, utilization, and reduces SLO violations under heavy load.

\noindent\textbf{Experimental setup.} We evaluated various models, which are summarized in Table~\ref{tab:model_overview}. To account for the different processing latency of each model, we adjusted the query per second (QPS) accordingly. The Twitter trace is mapped to a minimum of 5 QPS. We set the maximum QPS for each model based on its capacity to process requests within one second without missing any deadline. The maximum QPS for TVLT, AVHuBERT, TBN, MMSA, and ViLT is set to 60, 20, 40, 100, and 40, respectively. These values were $2\times$ the number of maximum requests each model is capable of handling under one second. We generated requests for each job following a normal distribution, with a mean of 1 and a standard deviation of 6, until the total number of requests matches the QPS. We randomly assigned the accuracy of each job within the range of the model’s performance, which was determined by the lowest and highest accuracy that the model can achieve using different modalities.

We used two types of approaches, modality-aware and modality-agnostic, to optimize the online performance of the models. We implemented four different policies for these approaches. The modality-aware approach includes: (a) \texttt{optimized}: described in Section~\ref{sec:online-section}, it attempts to utilize available resources to achieve the highest accuracy across all enqueued jobs; (b) \texttt{random}: described by Algorithm~\ref{alg:random_mod}, this policy randomly selects different jobs from the queue and applies the fastest strategy that meets the accuracy SLO. It repeats this process until no deadline violation can occur;(c) \texttt{aggressive}: different to the random policy, it selects \textit{all} enqueued jobs and applies the fastest strategy that satisfies the accuracy SLO. (d) The modality-agnostic policy (\texttt{none}) performs modality modification, and serves as the baseline.

\begin{figure*}[h]
    \centering
    \includegraphics[width=1\textwidth]{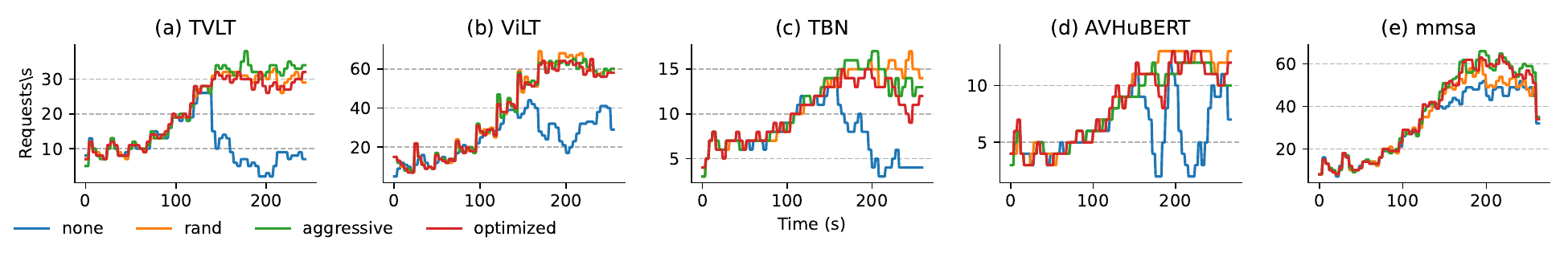}\hfill
    \includegraphics[width=1\textwidth]{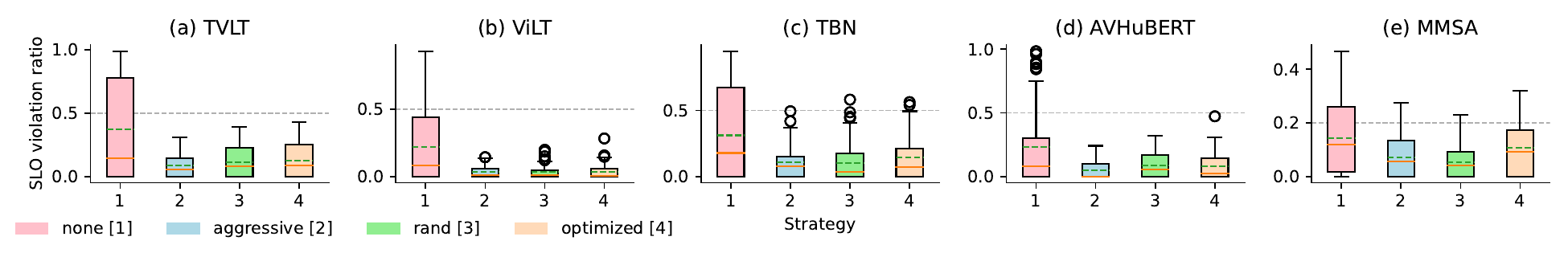}\hfill
    \caption{
    \textit{
    Throughput and SLO violation ratio (number of SLO violations by total number of requests), profiled every 4 seconds. Each box shows the outlier, median, mean, 25\%, and 75\% quartiles.
    }
    }
    \vspace{-1em} 
    \label{fig:on-time-ratio}
\end{figure*}

\noindent\textbf{Results and discussion.} Figure~\ref{fig:on-time-ratio} compares \proj{}, with dynamic modality selection. The throughput across all models is higher than the modality-agnostic method. TVLT, AvHUBERT, TBN, MMSA, and ViLT on average achieved $5.3\times$ $2.2\times$, $3.1\times$, $1.12\times$, and $4.3\times$ higher throughput, respectively. When the request arrival rate is low, both the modality-aware and modality-agnostic approaches have similar throughput. On the other hand, the modality-aware methods can handle higher request arrival rates, while the modality-agnostic method suffers from high processing latency and fluctuation. This is due to the high processing latency of the model. MMSA has much lower processing latency across all modalities, leading to close performance among different modality strategies.

\begin{figure}[h!]
    \centering
    \includegraphics[width=\linewidth]{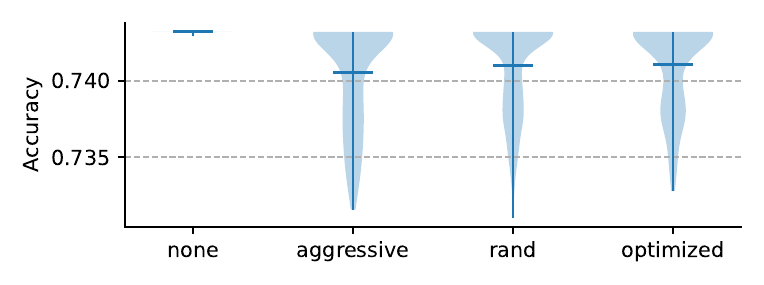}
    \caption{
    Accuracy distribution of TVLT with average accuracy of all jobs using different modality strategies.
    }
    \vspace{-1em} 
    \label{fig:trace-avg_acc}
\end{figure}

Figure~\ref{fig:on-time-ratio} also shows that all modality-aware techniques achieve much fewer SLO violations compared to the modality-agnostic approach. The optimized policy has
$25\%$, $18\%$, $17\%$, $15\%$, and $4\%$ lower average SLO violation ratio compared to the modality-agnostic approach for TVLT, VilT, TBN, AVHuBERT, MMSA, respectively. The optimized policy has in general a higher SLO violation ratio compared to the aggressive and random policy, due to the online optimizer overheads. It compensates for the slightly higher violation ratio by having higher average accuracy and more even accuracy distributions across jobs, as shown in Figure~\ref{fig:trace-avg_acc}.


\begin{figure*}[ht!]
    \centering
    \includegraphics[width=1\textwidth]{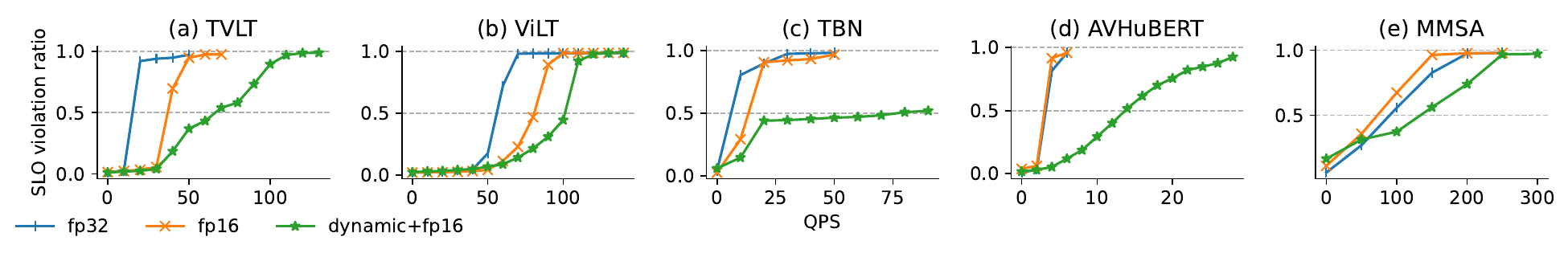}\hfill
    \caption{
    \textit{
    SLO violation ratio using FP32, FP16, and dynamic modality selection combined with FP16.
    }
    }
    \vspace{-1em}
    \label{fig:quant_violation_ratio}
\end{figure*}

\subsection{Complimenting Existing Approaches}
\label{subsec:compatibility}
Next, we show that \proj{} can be seamlessly incorporated into existing model optimization techniques to further improve inference throughput, addressing \textbf{}. 

\noindent\textbf{Experiment Setup.} We use quantization to show how modality-aware techniques can be combined with other model optimization techniques to further reduce inference latency and satisfy SLO objectives. Quantization is a technique that reduces the precision of numerical values in a model, from high-precision data types to low-precision data types~\cite{nagel2021white}. This technique can reduce memory footprint, as well as speed up the inference process. We use two data types for evaluation: \texttt{float32}, and \texttt{float16}. To study the effect under various loads, we select a range of QPS for each model. For modality selection, we use the \texttt{optimized} policy. The maximum QPS is set to the point where the deadline violation ratio reaches $99\%$.

\noindent\textbf{Results and discussion.} Figure~\ref{fig:quant_violation_ratio} shows that quantization allows all models to handle higher QPS before the deadline violation ratio rises sharply. For example, when using only quantization, AVHuBERT fails to increase its processing capability; when both quantization and dynamic modality selection are used, AVHuBERT can process up to $7\times$ more requests before reaching a $99\%$ violation ratio. This shows our approach is compatible with existing model optimization approaches and can improve their performances.


\subsection{\proj{}'s decision overheads}
\label{subsec:decision_overhead}
In the offline profiling stage, \proj{} performs two tasks: (a) it measures the latency of different modalities under various batch sizes, and (b) it generates the optimal modality selection strategies for each batch size. Table~\ref{tab:offline_stats} shows the median latency of these tasks and the speedup achieved by \proj{} over a brute-force search. Generating a single optimal modality offline selection strategy takes only 12ms.

\begin{table}[h]
  \centering
  \begin{tabularx}{\linewidth}{XXX}
  \hline
    \textbf{Profile(s)} & \textbf{Optimize(s)} & \textbf{Speedup} \\
    \hline
    32 & 45 & 31$\times$ \\
    \hline
  \end{tabularx}
  \caption{The amoutnt of time TVLT spends in both system metrics profiling and modality generations, as well as speedup compared to brute force search for optimal modality generations.}
  \label{tab:offline_stats}
\end{table}


In the online stage, \proj{} does two things: (a) it searches for the pre-computed optimal modality strategies that match the SLOs of each enqueued job, and (b) it finds the best modality selection strategy for each job. The optimizer’s overhead varies from 12 ms to 80 ms. Note that this not on the critical path on job execution, as we overlap the optimization process with the job execution by having enough jobs enqueued by worker, as discussed in Section~\ref{sec:implementation}.


\subsection{Ablation Study}
\label{subsec:ablation}
In this section, we show how variations in both offline and online optimization phases can affect the inference process.


\begin{figure}[h!]
    \centering
    \includegraphics[width=\linewidth]{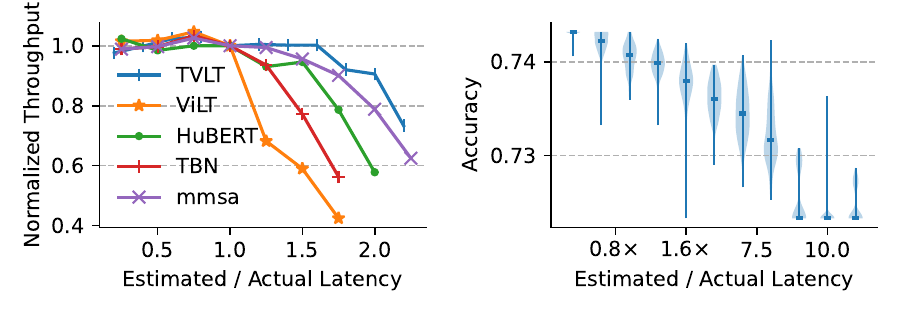}
    \caption{
   \textit{
    Left: demonstrating the effect of the deviation between the expected and actual execution latency on models' normalized throughput. The discrepancy is calculated by estimated latency over actual latency. Right: accuracy distribution under different discrepancy between estimated and actual execution latency for TVLT. The discrepancy is calculated by estimated latency over actual latency.
    }
    }
    \vspace{-1em}
    \label{fig:modality-mem-diff}
\end{figure}


\noindent\textbf{Experimental setup.} 
To evaluate the impact of the offline optimization, we first generate optimized modality selection strategies, discussed in Figure~\ref{sec:optimize-mod-selection}. We then vary the latency from $20\%$ to $250\%$ of the original latency, to simulate the potential discrepancy between the estimated latency and the actual inference latency on real hardware. To evaluate the impact on accuracy, we also use TVLT with fixed QPS of 40. We use \texttt{optimized} strategy for all experiments.

During the online phase, \proj{}'s overhead can vary significantly depending on how long it takes to finalize an optimized plan for all jobs in the queue. Therefore, we impose different time constraints on the online optimizer, ranging from $10$ ms to $120$ ms, to study how it affects the performance of \proj{}.

\noindent\textbf{Results and discussion.}
As Figure~\ref{fig:modality-mem-diff} shows, both throughout and accuracy can tolerate errors in latency estimation. All models can tolerate underestimated latency and maintain throughput. TVLT, AVHuBERT, and MMSA and tolerate up to $50\%$ latency overestimation with negligible sacrifice in throughput. Since it's rare to obverse such discrepancy in inference infrastructures~\cite{clockwork}, we believe \proj{} is robust against estimation errors in most scenarios. The changing accuracy, as shown in Figure~\ref{fig:modality-mem-diff}, is attributed to the system having false impression of resources due to overestimation, thus dropping jobs prematurely.




\section{Related Work}

\noindent\textbf{System-level dynamic optimization}
~\cite{clipper} proposes dynamic input batching to improve serving throughput by amortizing GPU kernel execution costs across multiple requests. It dynamically selects the largest profitable batch size that meets latency constraints. 

Serving systems dynamically assign GPUs to jobs based on their SLOs and request rates. Some of them ~\cite{nexus, SalusMLSys2020, gpulet} consider GPU sharing to improve GPU utilization and goodput. ~\cite{shepherd} proposes burst-tolerant resource provisioning by mapping multiple jobs to a group of resources at runtime.
~\cite{shepherd} argues that preemption is necessary to maximize a serving system’s goodput and their system makes preemption decisions at runtime providing formal guarantees on goodput.

~\cite{infaas} introduces a new dynamism layer, model-variants. A user specifies a task, accuracy, and latency requirements, and the proposed serving system automatically and dynamically explores the accuracy-latency tradeoff space of model-variants for the same task.  ~\cite{frugalgpt} generates cost-effective LLM cascade execution plans, leveraging different cost-accuracy characteristics of different LLMs.

~\cite{li2021low} focused on dealing with the delayed communication of input data in the case of multi-modal inference on streaming sensor data. Their proposed approach generates an input modality that is delayed based on the available input using a generative adversarial network (GAN) instead of waiting for the delayed input.  They assume that dropping a modality always causes a significant accuracy drop. 


\noindent\textbf{Model-level optimization}
A number of ML compression techniques~\cite{model-compression-survey} including pruning~\cite{xia2023flashllm,semi_sparsity} and quantization~\cite{nagel2021white} reduce both a model’s memory and computational costs by reducing model weights or precision. They are usually applied before deployment, but recent work shows that runtime quantization bit-width decision is beneficial~\cite{dynamic-quantization}.

Early exiting ~\cite{deebert, pabee, branchynet, berxit} adds task-specific layers (e.g. classification) to existing models, and stops inference early based on a given confidence level. ~\cite{draft-and-verify} uses layer skipping and output verification for LLMs. It dynamically skips layers to reduce per-token inference time.

In mixture-of-experts (MoE) models~\cite{moe-model}, a model is partially activated during its forward pass. A gating network selects the expert networks that will be activated based on input. This architecture allows a model’s parameters to scale while avoiding the prohibitive forward pass execution costs of a dense model with the same number of parameters.

Data multiplexing~\cite{datamux} adds multiplexing and demultiplexing layers at the beginning and end of the original model. The former transforms inputs into a succinct encoding and the latter does the opposite at the output.  This improves throughput as the original model only runs on the more succinct encoding space. This technique is complementary to our approach that {\em drops} portions of the input data. 


\section{Conclusions}

We present a new approach for dynamism, where we modulate the input to a model at inference time by selectively dropping portions of it. We should how the benefits of doing this in multi-modal inference. We highlight the key challenges that arise in leveraging this idea and present practical solutions to overcome them within our system, \proj{}.

We believe that input data modulation along with model optimization provides new possibilities for existing inference literature. The ability to modify the input data can lead to significant benefits across the entire inference serving stack, including reduced network bandwidth, lower preprocessing costs, energy efficiency, and reduced operating costs. We envision that \proj{} can be applied to a wide range of scenarios where input data variability is high and requires adaptive optimizations.

\bibliographystyle{plain}
\bibliography{main}
\end{document}